# Expediting data extraction using a large language model (LLM) and scoping review protocol: a methodological study within a complex scoping review

James Stewart-Evans [1,2]

Emma Wilson [1,3]

Tessa Langley [1]

Andrew Prayle [4, 5]

Angela Hands [6]

Karen Exley [2]

Jo Leonardi-Bee [1,3]

1. Nottingham Centre for Public Health and Epidemiology, University of Nottingham, Nottingham, UK
2. Environmental Hazards and Emergencies Department, UK Health Security Agency, UK
3. Centre for Evidence Based Healthcare*, University of Nottingham, Nottingham, UK
4. Lifespan and Population Health, University of Nottingham, Nottingham, UK
5. Nottingham Biomedical Research Centre, University of Nottingham, Nottingham, UK
6. Office for Health Improvement and Disparities, Department of Health and Social Care, UK

Corresponding author:

James Stewart-Evans

james.stewart-evans@nottingham.ac.uk

## Abstract

The data extraction stages of reviews are resource-intensive, and researchers may seek to expediate data extraction using online (large language models) LLMs and review protocols. Claude 3.5 Sonnet was used to trial two approaches that used a review protocol to prompt data extraction from 10 evidence sources included in a case study scoping review. A protocol-based approach was also used to review extracted data. Limited performance evaluation was undertaken which found high accuracy for the two extraction approaches (83.3% and 100%) when extracting simple, well-defined citation details; accuracy was lower (9.6% and 15.8%) when extracting more complex, subjective data items. Considering all data items, both approaches had precision >90% but low recall (<25%) and F1 scores (<40%). The context of a complex scoping review, open response types and methodological approach likely impacted performance due to missed and misattributed data. LLM feedback considered the baseline extraction accurate and suggested minor amendments: four of 15 (26.7%) to citation details and 8 of 38 (21.1%) to key findings data items were considered to potentially add value. However, when repeating the process with a dataset featuring deliberate errors, only 2 of 39 (5%) errors were detected. Review-protocol-based methods used for expediency require more robust performance evaluation across a range of LLMs and review contexts with comparison to conventional prompt engineering approaches. We recommend researchers evaluate and report LLM performance if using them similarly to conduct data extraction or review extracted data. LLM feedback contributed to protocol adaptation and may assist future review protocol drafting.

**Keywords**: data extraction; large language models; methodology; reviews; semi-automation

**Abstract word count:** 250

## Highlights

What is already known?

- LLMs are increasingly used as a tool in reviews, and automation of searches and screening of evidence sources are common applications
- Issues noted with respect to performance of data extraction by LLMs include difficulties handling nuanced, subjective criteria; a lack of analytical depth and attention to detail; inaccuracies due to reliance on pre-trained knowledge; and LLM "hallucinations"
- Whilst convenience, facilitation and timesaving are rationales for LLM use, their conventional

tasking is through iterative prompt engineering which seeks to optimize performance

What is new?

- This methodological study investigated the use of a large language model (LLM), Claude 3.5 Sonnet, to expedite (rather than optimize) data extraction for a scoping review
- Two LLM data extraction approaches were evaluated, which used a scoping review protocol to prompt data extraction rather than conventional prompt engineering
- Extraction performance was notably higher for simple, well-defined data items in comparison to more complex and subjective data items
- The LLM was also used to verify data that had been manually extracted by a human reviewer: LLM feedback added value but LLM detection of deliberate data extraction errors was low

Potential impact for readers

- Review protocols and data extraction instruments can be improved using LLM feedback and used to specify LLM tasks and expedite data extraction, but performance may be substandard
- Performance considerations include evidence source heterogenicity, data item definition, and process granularity (individual or batch processing of evidence sources or data items)
- Evaluation and reporting of feedback validity is advisable if using LLMs to verify data extracted by human reviewers

## Introduction

Due to the broad nature of scoping reviews, it is common for a large number of included sources to undergo data extraction [1]. Data extraction instruments for scoping reviews may specify multiple or broadly defined data items, allowing for extensive free-text excerpts to be extracted. Data extraction tends to be resource intensive when scoping reviews display all these characteristics, which is a particular challenge for smaller review teams. Large language models (LLMs) - machine learning models trained on human language that aim to predict and generate plausible language in response to an input [2] - offer the possibility of automating some manual aspects of data extraction to reduce the burden on human reviewers.

The potential use of tools and software to manage screening and support the selection of evidence sources and their data extraction is acknowledged by methodological guidance such as the *JBI Manual* *for Evidence Synthesis* [3], wherein data extraction tools are document templates. The state of science in data extraction automation in systematic reviews was summarized by a systematic review in 2015 [4,5] and subsequently in a living systematic review (last updated in 2025) [6]. The reviews found publications focused on a limited number of PICO-orientated (Population, Intervention, Comparison, Outcome) data items and that Natural Language Processing (NLP) techniques (which enable computers to process human language [7]) were not fully used to automate data extraction. Though the quality of studies' reporting was noted as becoming more complete and clearer up to 2021 [6], there was no standard approach to performance evaluation. Requirements for training data and difficulties extracting data from portable document format (PDF) documents hampered past use of NLP models [8]. Whilst data extraction remains largely a manual process involving reading and electronic searches of text [5,9], a proliferation of LLMs have subsequently become widely available to researchers [6,8-10].

Emergent abilities of LLMs include other tasks such as summarization and classification of text [2]. Studies and commentaries have begun to explore the use of artificial intelligence (AI), such as LLMs, as a tool in reviews or replacement for humans in the review and evidence synthesis process [10], potentially integrating specialist NLP techniques eg, [11,12]. As LLMs are multifunctional, they can potentially assist with multiple tasks within a review [9,10,12-15], though fewer automation projects have focused on the full cycle of review automation [10,12]. Automation of searches and screening of evidence sources are of specific interest, the latter being well addressed by existing tools [6], including some review-specific LLMs. Most relevant to this paper is an area of emerging interest: the performance of data extraction by LLMs [8-10,16-18]. Potential issues noted in the literature include difficulties handling nuanced, subjective criteria; a lack of analytical depth and attention to detail; incorrect citations; and inaccuracies due to reliance on pre-trained knowledge and LLM "hallucinations": answers which appear plausible but are incorrect [6,7,10,13]. These issues are compounded by a reported trend of lower reproducibility of results and decreased quality of reporting in comparison to previous data extraction studies; evaluation is challenged by the need for human evaluation and LLM performance that varies between prompt and LLM updates [6].

The aim of this study was to trial the use of an LLM to expedite aspects of data extraction during the conduct of a complex scoping review entitled *Health net-outcome objectives and approaches for spatial planning and development: a scoping review* [19,20], which followed the JBI methodology [21]. Expedition was based on the premise that LLMs require no programming skills or prior curation of training data [6], and solo researchers may seek to use LLMs to save time during complex reviews without prior expertise in prompt engineering. As such, the focus was to explore proof-of-concept of an intentionally naïve approach adapting existing review-specific protocols and guidance as an alternative to conventional iterative prompt engineering (i.e., the aim was not to optimize extraction performance).

Evidence sources included in the case study review were heterogenous: they comprised academic and grey literature describing various net-outcome policy objectives relevant to health in spatial planning from different disciplinary perspectives and contexts of policy and practice (see Appendix V). LLM methods were developed and trialed in autumn 2024 after the completion of baseline data extraction from 119 evidence sources by a single human reviewer. Extracted data items included simple, well-defined items (such as citation details) and more complex and subjective data items. The latter included characteristics of policies and contextual settings, including strengths and weaknesses and domain and context-specific opportunities and challenges. Data extraction for these items was open-ended, requiring extraction of any relevant text excerpts from any part of evidence sources.

## Methods

### *Sample, baseline and LLM selection*

A randomized sample was taken of 10% (n=12 out of 119) of evidence sources included in the case study review [20] by selecting sources from a list created using sorted random numbers in Excel. Two evidence sources were subsequently excluded due to copyright considerations (n=10). The sample comprised PDFs written in English language and included the scoping review protocol [19], which was an included evidence source. This inspired its use in this study as both an evidence source and an LLM task prompt.

The original data extraction for the sample of 10 evidence sources was used as a baseline for LLM data extraction. Post-hoc checks of the original extraction were conducted by a second human reviewer and the LLM itself (reported later in this paper). No significant data extraction issues or errors were identified.

Claude 3.5 Sonnet (Anthropic, CA, US) was selected to i) conduct expediated data extraction and ii) review data extracted by a human reviewer. Claude 3.5 Sonnet was described by its developer, Anthropic, as the most intelligent model of LLMs in the Claude 3.5 family; their stated capabilities include reasoning, coding, multilingual tasks, long-context handling, honesty, and image processing [22]. Claude was, at the time of method development, one of few well-developed generalist LLMs with a longer context length, enabling it to handle detailed information and process data extraction instructions and multiple evidence sources (i.e. multiple documents) within a single prompt, which was a requirement of the methodological approach. Its capacity to understand complex queries and contextual relationships and adapt to non-standard formats and reporting were also considered relevant to the context of the case study review.

## *Method development*

Prior to the evaluation of protocol-based data extraction prompts, user-LLM interactions explored approaches and trialed aspects of the final methodology. Performance was not formally evaluated; this stage was used to develop executable prompts and associated reference materials and comprised:

- User definition of net-outcome objectives and related terms and definition of selected data items (principles, strengths, weaknesses, opportunities, and threats) followed by item-by-item prompts to extract data from the scoping review protocol itself
- Aggregation of prompts into a series of all-in-one text prompts providing background information and task instructions. Tasks initially focused on extracting multiple data items from an exemplar extraction csv file (see next section), then extraction from individual and batched evidence source PDFs into csv format. Multiple rounds of user-requested LLM feedback were used to improve prompt specification, which included adaptation of the protocol data extraction instrument. Corrective prompts reiterated adherence to prompt instructions and extraction of evidence source content not generative synthesized outputs
- Creation of a project workspace in which instructions and an exemplar dataset were uploaded and then referenced in a prompt to extract data from the scoping review protocol itself. Corrective prompts reiterated adherence to uploaded instructions and extraction of only evidence source content until output format and content were judged compliant

Thereafter, two variants of data extraction instructions were developed which used the review protocol as a prompt to expediate data extraction. Earlier conversation histories were retained, as this may improve LLM learning and performance by retaining the dialogue context available to an LLM when it is producing a new response [15].

## *Protocol-based LLM data extraction*

Extended protocol approach (development)

A document package was compiled comparable to materials provided to a second human reviewer prior to independent extraction. Documents comprised the scoping review protocol [19], an adapted data extraction instrument containing categorizations and elaborated descriptions of data items and data extraction conventions (Appendix II), step-by-step data extraction and presentation instructions informed by earlier method development (Appendix III), and a csv file containing examples of data extraction. Data extraction targeted three simple, well-defined data items (citation details: author(s), publication year, title) and five more complex and subjective data items (key findings: implementation principles, strengths, weaknesses, opportunities, threats).

Examples of data extraction comprised eight publicly available evidence sources of a new 10% sample (n = 12 out of 119) of evidence sources included in the scoping review, ordered alphabetically by author surnames. These examples did not include any evidence sources selected for LLM data extraction.

Extended protocol approach (application)

An 'extended protocol' approach was trialed within a project workspace in which the document package was first uploaded. The 10 evidence sources selected for LLM data extraction were then individually processed in alphabetical order of author surnames. Each evidence source was uploaded to the user interface in PDF format with a text data extraction prompt (Appendix I) referring to the uploaded reference documents. The scoping review protocol – as it was also an eligible evidence source – was last to be processed. Dialogue continued until the interface prohibited further prompts or uploads, at which point the process resumed in a new user-LLM conversation.

Dynamic elements of user feedback and corrective prompts were included. Additional prompts were provided after considering the format and content of extracted data with reference to LLM instructions and evidence source content. The process sought to correct any obvious data extraction omissions or errors or slippage of content or format of LLM outputs and is summarized in Appendix I.

Simple protocol approach

To trial an unrefined approach, a simple 'protocol' data extraction task prompt was used in which each evidence source was uploaded to the default user interface in PDF format along with the scoping review protocol [19] and a text data extraction prompt (Appendix I). No examples of data extraction were provided (i.e., this was a form of zero-shot prompt).

Whilst the extended protocol approach specified the policy objective described in each evidence source to assist targeted data extraction, in the simple approach the focal objective was unspecified. Furthermore, no constraints were applied (i.e., the LLM output all data items).

An evidence source familiar to the human reviewer [23] was processed first to check for any obvious data extraction issues; evidence sources were then processed in alphabetical order of author surname. A modified version of the initial task prompt (Appendix I) was used for the second and subsequent evidence sources to emphasize extraction of data only from the evidence source.

Analysis of protocol-based LLM data extraction

Extracted data was tabulated in Excel alongside the baseline extraction. In cases featuring corrective prompts, LLM responses to the final prompt were used. The extracted net-outcome objective was recorded for the simple protocol approach. Otherwise, for both methods, tabulated data items comprised the eight data items targeted by the extended protocol approach as previously described. Key findings cells contained multiple lines of text excerpts. Each excerpt line was classified by the reviewer who had carried out the baseline data extraction as either:

- relevant (fully or partially overlapping with or similar in meaning to baseline excerpts, or – if a new excerpt – within the data item scope), or
- misclassified (relevant to another data item), or
- irrelevant (not relevant or misclassified and not within the data item scope).

Excerpt classification frequency counts were compiled for each evidence source. If excerpts of the first two types were new (not present in the baseline), this was noted. Ineligible excerpts (e.g., data from other evidence sources) were noted.

Secondly, citation data items were compared to the baseline extraction and categorized by the reviewer who had carried out the baseline data extraction as true or false positives or negatives [8]. Key findings cells contained multiple text excerpts, which challenges definition of true negative entities [6]. In this case, each excerpt was individually categorized (as a true or false positive or negative). True positives were attributed when extracted data contained equivalent information or excerpts matching the baseline (permitting inconsistencies in formatting or truncation). If several short excerpts overlapped with a longer baseline excerpt, this was considered a single true positive (or single false negative, if incorrectly attributed to another data item). False positives were identified by checking evidence source content and the complete baseline extraction.

Results were tabulated as frequency counts of true positives (TP), true negatives (TN), false positives (FP), and false negatives (FN) for each data item for each evidence source. Accuracy, precision, recall, and F1 scores (a combined measure of precision and recall) [6,8-10] were calculated as outlined below). Gartlehner et al. [8] provide further elaboration of relevant terms and types of error.

- Accuracy: (TP+TN)/(TP+TN+FP+FN)
- Precision (P) (also known as positive predictive value): TP/(TP+FP)
- Recall (R) (also known as sensitivity): TP/(TP+FN)
- F1 score: 2*(P *R)/(P+R)

Summary metrics were calculated in two ways: a micro metric was calculated using frequency counts (i.e., total TPs, TNs etc.), and a macro metric was calculated as an average across the eight data items (i.e., giving equal weight to each item regardless of the number of excerpts).

Lastly, all excerpts were listed for the evidence source with the most baseline excerpts, and the status of each (LLM-extracted: yes/no, and if yes the excerpt classification as above) was noted along with any reviewer observations.

*Protocol-based LLM review of baseline data extraction*

Using a single task prompt within the project workspace (Appendix IV, batches 1-2), the LLM was tasked to review a new csv file containing the baseline extraction of the 10 evidence sources previously used in LLM data extraction. The task prompt referenced evidence sources (uploaded in two batches of five PDFs), the scoping review protocol [19], and the original data extraction instrument (Appendix V) used for the baseline extraction. Each batch was processed in a new conversation. The baseline extraction included most of the original scoping review data items and a new field specifying evidence source filenames.

Review prompts were then updated (Appendix IV, batches 3-4) and deliberate errors were introduced to the baseline data extraction. The modified prompt emphasized careful comparison of the baseline extraction to evidence source content. Deliberate errors comprised publication dates, types of policy objective described, misattribution of data items (swapping of column content), misattribution of sources (swapping of row content), and random sentence excerpts. The review exercise was repeated in a new series of conversations within the project workspace.

Analysis of protocol-based LLM review of baseline data extraction

LLM feedback was tabulated in Excel, differentiating between source citation and key findings data items. Frequency counts of proposed revisions were calculated, as was the proportion considered to add value by a reviewer who had carried out baseline data extraction; these comprised LLM corrections of true errors and discretionary suggestions. Any LLM feedback that contained ineligible excerpts (e.g., information extracted from another source) was noted.

Detection of deliberate errors was recorded as a yes/no response, noting verbatim LLM feedback when given. Detection was summarized per evidence source and per error type.

## Results

### *Protocol-based LLM data extraction performance*

Table 1 presents data extraction performance for the two approaches (protocol and extended protocol) compared to the baseline extraction. Excerpt totals relate to key findings data items (implementation principles, strengths, weaknesses, opportunities, and threats; see Appendix II) that contained multiple excerpts.

<placeholder for Table 1: LLM data extraction performance versus a human reviewer (performance per evidence source)>

*Denotes that evidence source #10 was also the scoping review protocol.

Both methods generated fewer relevant excerpts (protocol: n=26; extended protocol: n=47) than the human reviewer (n=206). In comparison with the baseline, LLM excerpts were generally shorter text strings; truncation occurred when context was excluded and long sentences were split into independent points. Unlike the baseline extraction, LLM excerpts were not listed by order of appearance in source text.

Compared to the protocol method, the extended protocol method returned a higher proportion of relevant excerpts (45.2% versus 33.8%) and a lower proportion of irrelevant excerpts (22.1% versus 35.1%). Both methods returned a similar proportion of misclassified excerpts (32.7%, versus 31.2%), and the protocol method returned slightly more new excerpts (n=13 versus n=10).

There were qualitative differences in extracted data between the two methods, which returned common and differing excerpts. Both presented some excerpts that did not originate from the specified evidence source (protocol: n=1; extended protocol: n=2). Both methods' performance varied between evidence sources in comparison to the baseline in terms of numbers of relevant excerpts and new excerpts. For the extended protocol approach, user feedback (Appendix I) appeared not to materially improve subsequent performance.

Table 2 presents data extraction performance metrics; this section summarizes overall micro (rather than macro) scores [6]. The simple protocol (83.3%) and extended protocol (100%) approaches displayed high levels of accuracy when extracting citation details; accuracy was much lower (9.6% and 15.8% respectively) when extracting data items related to key findings. Key findings data items had comparatively low counts of true positives and high counts of false negatives comprising both missed excerpts and misattributed excerpts. For citation detail data items, both approaches had high (100%) precision as no false positives were returned, whilst the recall of the protocol approach was comparatively lower (83.3% versus 100%) because the LLM extracted no title for five of 10 evidence sources. Overall, the extended protocol approach showed higher levels of precision and recall and higher F1 scores in comparison to the protocol approach. Considering all data items, both approaches had precision >90% but low recall (<25%) and F1 scores (<40%).

*<placeholder for Table 2: LLM data extraction performance versus a human reviewer (performance per data item)>*

*Legend: Micro averages are calculated across all document excerpts; macro averages are calculated across data items.*

Considering the LLM-defined policy objective in the simple protocol approach, of the 10 evidence sources, the extracted content identified a relevant objective in four cases, identified a valid related alternative focus in two cases, and focused on another objective in the four remaining cases.

In terms of excerpts from the evidence source with the most baseline excerpts (source #9, Table 2),

of 24 baseline excerpts also extracted by one or the other LLM approach, only one was extracted by both LLM approaches and 23 were unique. Both approaches returned excerpts whose content included data item title words (notably "principle" and "challenge"), though these words were also present in baseline excerpts missed by the LLM. Key findings data item excerpts returned by the extended protocol approach contained references to the user-specified policy objective, HNG (health net gain), or, less frequently, more general references to net gain(s) or determinants of health. Two baseline excerpts spanned multiple PDF columns or pages: these were extracted by one or the other approach (i.e., were not missed).

*Protocol-based LLM review performance*

When asked to review the scoping review's data extraction guidelines, the LLM stated that no significant issues were present and that the guidelines "…appear to be comprehensive and well-structured, allowing for a thorough and consistent extraction of relevant information across different types of sources." Five minor points that could be clarified or refined were suggested.

LLM summary feedback on baseline extractions varied in exact wording but ubiquitously stated that extractions were accurate, with some areas identified for improvement. Table 3 presents a summary. Overall, the LLM proposed 15 corrections to citation details, of which four (26.7%) were considered to potentially add value. These comprised a more detailed publication date, synthesized suggestions to populate keywords unspecified by two evidence sources, and a typographic correction. Discounted corrections related to duplicates (e.g., revised publication dates identical to the baseline), revised referencing styles unspecified by the data extraction instrument, and discretional typographical revisions.

<placeholder for Table 3: LLM data extraction review performance>

For key findings data items, the LLM proposed 38 excerpts for inclusion, of which 8 (21.1%) were considered to potentially add value. These comprised general and specific suggestions to elaborate existing entries (in five cases providing suggested additional excerpts). Discounted suggestions related primarily to duplicates. The LLM also proposed four ineligible excerpts (hallucinations) for three evidence sources: these originated from other evidence sources (on two occasions) and the baseline extraction (on two occasions).

Overall, LLM review identified no major errors in human data extraction and few minor errors (as defined by Gartlehner et al. [8]). There were, however, differences in output and performance between the first and second batches of processed evidence sources (Table 3), despite user prompts to maintain an identical approach. For the second batch, the LLM returned feedback on every data extraction item for each evidence source rather than LLM-selected data items, presented fewer but more relevant corrections, and included no ineligible excerpts.

LLM detection of deliberately introduced errors

*<placeholder for Table 4: LLM data extraction review performance (after introducing deliberate errors in human data extraction)>*

Table 4 summarizes LLM feedback on a version of the baseline data extraction that included deliberate errors. The LLM asserted that incorrect publication dates were correctly extracted for nine evidence sources but correctly reported one of ten as an error. An identical objective type (health net gain) had been listed for every evidence source; the LLM incorrectly described this as correctly extracted for each of the seven sources that described another type of objective. The swapping of two columns' entries (misattribution of data items) was undetected, as was the swapping of rows (misattribution of evidence sources). As these two error types affected four data items per evidence source, 40 cells containing incorrectly extracted data were not identified. Originating page numbers were cited in feedback reporting correct extraction of these cells. Of four inserted random text excerpts, the LLM correctly identified one as unrelated to evidence source content. Cells containing the three others were reported to be correctly extracted and specific evidence source page numbers were spuriously cited as the origin in two cases.

## Discussion

The novelty of this methodological study relates mainly to data extraction challenges posed by the scoping review context, our use of a scoping review protocol [19] and data extraction instrument as task prompts to expedite data extraction, and our use of an LLM as a second reviewer of a baseline data extraction. We focus our discussion on each aspect.

### *Considering review context during evaluations*

Our findings confirmed the importance of considering review and data item types when using or evaluating LLMs. A focus of studies of assisted data extraction on PICO data items has previously been

*Page 13*

reported [4-6], which is continued in LLM studies [8,9,11,16]. These may often focus on extracting well-defined numeric or text data from sources with well-defined study designs and standard reporting.

The characteristics of scoping reviews – namely complexity and heterogeneity of evidence sources and exploration of concepts and boundaries [24] – may influence data extraction performance. Domain adaptation is a challenge for data extraction systems [6], including LLMs [9]. Whilst pre-trained domain-specific LLMs display improved performance [13], included evidence sources may span disciplines. Scoping reviews have infrequently (in <5% of cases) been the focus of past LLM automation projects [10], and, in contrast to others [6,8,9,16], our case study was not a systematic review of clinical or epidemiological studies, interventions, or structured data reported as primary outcomes. Direct comparisons of LLM performance are hindered by, inter alia, the use of different datasets which influence the difficulty of data extraction [6]. Selection of quantitative and qualitative performance metrics in different review contexts and data extraction performance across different types of review merit further research, as do opportunities for standardization [6,9].

*Protocol-based data extraction*

We used two LLM approaches to extract data and demonstrated that a scoping review protocol and supporting materials could be used to specify LLM data extraction tasks and expedite data extraction with relatively little effort. We used LLM feedback on document content to develop an extended protocol approach, and similar approaches may assist researchers with the iterative development of review protocols and data extraction instruments (regardless of an LLM's role in data extraction).

Performance is affected by differing approaches to data extraction and parameterization, and LLM extractions have typically constituted generative summaries of data of interest [6]. In contrast, our case study required exhaustive verbatim extraction, and the overall performance of data extraction was substandard. Accuracy can vary between LLMs presented with identical tasks [16], and automation projects have used various LLMs and performance metrics, reporting typically higher levels of data extraction accuracy (e.g., >80%) than was found in our trial (<25% considering all data items) [6,8,10,16]. Our recall and F1 scores were similarly within the lower end of reported ranges or would constitute outliners. It is important to disaggregate data items within overall performance metrics. Whilst the LLM in our trial displayed high accuracy when extracting simple, well-defined citation details (Table 2, 83-100%), accuracy was much lower when extracting more complex and subjective key findings data items (Table 2, 10-16%). These items required semantic and contextual interpretation, and poor performance related primarily to misattributed and missing excerpts. An evaluation of GPT-4 similarly reported a tendency towards over-exclusion through false negatives [17]. We noted also that LLM excerpts displayed shorter text string lengths than baseline excerpts. Other evaluations have similarly reported higher accuracy for smaller, simpler types of data, and binary and Boolean data types in comparison to strings (open response types) [9].

Task prompt specification is a related consideration [9,17,25]. Direct processing of text, rather than individual PDFs, may improve performance [8,16]. Whilst we found an extended protocol approach was more successful at extracting key findings, performance remained poor. Ambiguity is a challenge for LLMs when defining data items and interpreting evidence source content [6], and the location and nature of key findings data items was less well defined than citation details in our study. Our extended protocol featured some limited fine-tuning and an example data extraction dataset and approach similar to "few-shot prompting" [26]; this could have been further developed through inclusion of originating evidence sources. Conventional iterative prompt engineering approaches focused on optimizing extraction of each individual data item also seem well founded if using LLMs to extract items that require extensive discussion, training or guidance for human reviewers. LLM relation extraction tasks remain problematic for LLMs [6], and conditional relationships between data items may also be important to explicate (such as, in our case, an identified policy objective linked to key findings). We observed that more relevant data was extracted when evidence sources described a single objective or when a focal objective was user-defined.

We found corrective prompts were also necessary to correct slippages from user-specified output formats. Although LLMs are designed to respond using predictions rather than retrieval [27], the LLM in our trial generally adhered to instructions to extract verbatim text, with some false positives. Validation of outputs was time-consuming but remains important if using an expediated approach. If expediency is the priority, a simple protocol approach can be taken, and higher performance seems likely if dealing with standardized evidence sources and well-defined data items. Our study found that developing an extended protocol approach was time-consuming and did not sufficiently improve data extraction performance. Trials of instruction sheets have similarly reported mixed success [18]. Overall, our findings suggest that protocol-based data extraction might only be suitable to provide an initial extraction for human reviewers to check and extend. Other authors recommend similarly caveated use of LLMs for semi-automation [6-9,14].

## LLM review performance

When tasked to review baseline data extraction using batched evidence sources and reference documents including the review protocol, LLM feedback included some useful suggestions (including typographic corrections, elaboration of short or missing entries, and some relevant new excerpts). LLM commentaries sometimes reinforced contextual considerations previously noted by the human reviewers and provided a perspective on the inclusion or exclusion of borderline excerpts. Our experience suggests that human reviewers may find similar approaches useful to solicit LLM feedback on specific, contested items. In terms of overall review performance, however, the LLM proposed many duplicate entries already present in the baseline and some ineligible excerpts. Few deliberately-introduced data extraction errors were detected. This implies limited and inconsistent comparison of evidence source content to the baseline.

Further to factors previously discussed, Tao et al. [18] note that LLMs process limited amounts of text at once, which affects cross-referencing of details in longer documents. Memory limitations are also associated with deviations or inaccuracies in understanding of user prompts [15]. We used single prompts to process multiple tasks and documents. Although we specifically selected an LLM with a long content length, our multi-document protocol-based approach to extraction and batch approach to review may have influenced performance.

LLMs could foreseeably be used by researchers to review extracted data with reference to review protocols and evidence sources. In our case study, however, the LLM was an unreliable reviewer. Given the verisimilitude of LLM feedback, which included confident statements referencing originating page numbers, we recommend that researchers evaluate and report on LLM review validity. Other authors emphasize the importance of interpretability, explainability, and evaluation when faced with inscrutable internal reasoning and task execution [6,9,17]. Newer "reasoning" models [28] may help researchers to elaborate or explicate steps taken.

## Limitations

We used an intentionally naïve protocol-based approach to expediate data extraction and review as an alternative to conventional prompt engineering. The aim was to explore proof-of-concept not to optimize and extensively evaluate extraction performance, which, as previously discussed, was fundamentally challenged by the context of our case study scoping review and methodological approach.

Our trial was also limited by the limited number of evidence sources included (n=10) and tasking of a single LLM. Our extended protocol approach could have further elaborated data item definitions and data extraction heuristics to improve data extraction performance. We did not separate training and testing data when developing the methodology (i.e., prompts were developed using evidence sources also used in the performance evaluation), which is reported as commonplace and undesirable [6,9]. Data extraction and review performance may have been influenced by our use of publicly available evidence sources already included in the LLM's training dataset [8,9,16], prompt [9] and source processing order, and reflexivity across user-LLM interactions [18]. In common with some similar studies, our baseline extraction contained few instances in which data items were not reported, which affects the overall scope of evaluation [8].

Rigorous performance evaluation would entail direct comparison of review protocol-based and conventional prompt engineering approaches across a range of review contexts and LLMs. Our trial was carried out during the conduct of a review in a protocoled fashion, rather than according to a *priori* Study Within A Review (SWAR) protocol. It focused on exploring proof-of-concept within the scope and timelines of existing work and limited resources were available for method development or comprehensive performance evaluation. The reference standard was a single reviewer extraction. This was checked by a second human reviewer and found to contain no significant data extraction issues or errors, and we consider it unlikely to have significantly influenced performance metrics given that poor performance related to missed and misattributed excerpts (which imply limitations of the methodological approach and case study review context). However, the LLM performance evaluation itself was conducted by the reviewer who had carried out the baseline data extraction, not in a dual and independent manner. Future performance evaluations require dual evaluation and a double data extraction baseline to address potential performance biases, accepting that human-led data extraction remains an imperfect reference standard regardless of the use of dual-workflows [6,8]. More extensive evaluation is also necessary if protocol-based approaches are to be thoroughly evaluated, noting that we did not evaluate test-retest reliability or conduct sensitivity analysis in our limited trial.

## Conclusions

Researchers can use pre-existing review protocols and data extraction instruments to specify LLM data extraction tasks, and LLM input to the protocol drafting stage of reviews and role of review protocols in LLM data extraction deserve further attention. Our case study scoping review trial found one LLM's data extraction performance varied between evidence sources and two review-protocol-based methodological approaches and that though ours was a subpar substitute, LLMs might add value if used to extract provisional data prior to human researchers' data extraction. When used as a second reviewer of extracted data in our trial, the LLM added some value but was an unreliable arbiter of whether data had been extracted correctly. We recommend that researchers evaluate and report on LLM performance and feedback validity if using them similarly within reviews and propose future research might characterize practice and evaluation study findings; develop standardized LLM guidance, data extraction methodologies and templates; and formally evaluate and compare LLMs' use as assistants during the conduct of different types of review.

## Appendix I: LLM data extraction prompts

Note to readers: emboldened text in later subsections denotes iterative additions and amendments to earlier prompts.

### *Extended protocol LLM prompt*

Your role is to carry out data extraction from source documents that will be uploaded as PDF files into the chat by the user. The bullet points below are supporting information that lists and describes project documents and provides further instructions; please consider each project document and instruction in order:

- "Scoping review protocol.pdf" – this scoping review protocol project document describes a scoping review protocol. Use it to define the concept and scope of relevant net outcome objectives and their associated approaches.
- "Data extraction instrument (LLM).docx" – this data extraction instrument project document contains a data extraction template. Use it to identify data extraction targets (listed in the left hand column) in new source documents (which will be uploaded as PDF files into the chat by the user), strictly adhering to any guidance on the concept and scope of each data extraction target (in the right hand column).
- "Data extraction and presentation instructions.docx" – this document provides step-by-step data extraction and presentation instructions that must be strictly adhered to after new source document PDFs are uploaded into the chat by the user.
- "Data extraction examples.csv" – this document contains completed examples of data extraction that applied the scoping review protocol and data extraction instrument project documents to source document PDFs: it is provided for reference only and should not be extracted or included in any output.

[Prompt accompanying each evidence source PDF subsequently uploaded for processing] This document references a [user-specified, based on document content] net-outcome objective.

### *Extended protocol LLM corrective prompt summary*

For the first processed evidence source, user-LLM feedback comprised corrective prompts to: 1) adhere to the original task prompt, 2) note specified data extraction exclusions, 3) consider two differentiated policy objectives synonymous, and 4) amend formatting.

Corrections for the second evidence source comprised prompts to: 1) adhere to the original task prompt, 2) note specified data extraction exclusions, and 3) note specified data extraction inclusions.

For the third evidence source they comprised: 1) a prompt to amend formatting, 2) user feedback on all prior data extractions, stating that citation details were appropriate but that extracted key findings included misclassifications and omissions, and 3) a prompt to only extract text from the evidence source.

For the seventh evidence source they comprised a prompt to only extract text from the evidence source.

For the eighth evidence source they comprised a prompt to correct output formatting.

No corrective prompts were judged necessary for the fourth to sixth and ninth to tenth processed evidence sources.

## *Simple protocol LLM prompt*

### *V1.0 (first of ten evidence sources)*

Your role is to assist with data extraction. Apply the scoping review protocol to document X.pdf and present data extraction targets as headers with bullet point lists of relevant excerpts from X.pdf

### *V1.1 (second to tenth of ten evidence sources)*

Your role is to assist with data extraction. **Carefully read the scoping review protocol and apply it to document X.pdf, presenting extracted data as a series of** headers with bullet point lists of relevant excerpts from X.pdf. Only extract data from X.pdf, **not from the scoping review protocol itself**.

# Appendix II: Data extraction instrument (LLM)

| Evidence source details | |
|---|---|
| Author(s) | (Use Vancouver referencing style) |
| Publication year | |
| Title | |
| **Net outcome objective and its associated approach: characteristics and contextual positioning** | |
| Implementation principles | "Implementation principles" of a net outcome objective and its associated approach encompass: <br><br>• Principles, rules or requirements that govern implementation <br>• Items framed as pre-requisites, absolutes, essentials, necessary steps, imperatives, or propositions <br>• Implementation frameworks, "how to" guides, "steps to take", and action lists <br>• Descriptions of how a net outcome objective is implemented in practice or would be implemented in practice <br>• Descriptions of how a net outcome objective works or operates in practice or how it would work or operate in practice <br>• Descriptions of how a net outcome objective is or would be operationalised <br><br>If a given implementation principle is framed as a possibility or one of a range of options, it is within the concept of "opportunities" (and should be extracted under "opportunities" rather than "implementation principles"). <br><br>"Implementation principles" of a net outcome objective and its associated approach generally exclude: <br><br>• Items that are not clearly linked to net outcome objectives or their associated approaches (e.g., excluding implementation principles |
| | related to broader / multi-outcome objectives, such as regenerative design and development, unless explicit net principles are specified)<br>• Items unambiguously unrelated or untransferable to health, community, social, broad environmental, or generic net outcome contexts (e.g., text addressing implementation principles related to standalone ecological or resource-orientated objectives) |
| Strengths | "Strengths" of a net outcome objective and its associated approach:<br><br>• Are internal factors (i.e. inherent to the net outcome objective and its associated approach)<br><br>• Are positive effects or implications<br><br>• Include observed, demonstrated, or proven positive effects or implications<br><br>• Include advantages versus the status quo or an existing comparator<br><br>"Strengths" of a net outcome objective and its associated approach generally exclude:<br><br>• Items that are not clearly linked to net outcome objectives or their associated approaches<br>• Items unambiguously unrelated or untransferable to health, community, social, broad environmental, or generic net outcome contexts (e.g., text addressing strengths related to standalone ecological or resource-orientated objectives)<br><br>"Unstated" should be recorded if the concept is not addressed within the source. "Aggregated (not extracted)" should be recorded if the source presents content not specific to net-outcome aspects of an objective or approach. |
| Weaknesses | "Weaknesses" of a net outcome objective and its associated approach:<br><br>• Are internal factors (i.e. inherent to the net outcome objective and its associated approach)<br><br>• Are negative effects or implications |
| | - Include observed, demonstrated, or proven negative effects or implications<br>- Include disadvantages versus the status quo or an existing comparator<br><br>"Weaknesses" of a net outcome objective and its associated approach generally exclude:<br><br>- Items that are not clearly linked to net outcome objectives or their associated approaches<br>- Items unambiguously unrelated or untransferable to health, community, social, broad environmental, or generic net outcome contexts (e.g., text addressing weaknesses related to standalone ecological or resource-orientated objectives)<br><br>"Unstated" should be recorded if the concept is not addressed within the source. "Aggregated (not extracted)" should be recorded if the source presents content not specific to net-outcome aspects of an objective or approach. |
| Opportunities | "Opportunities" that apply to a net outcome objective and its associated approach:<br><br>- Are external factors (i.e. not inherent to the net outcome objective and its associated approach)<br>- Are context-specific and relate to its contextual positioning<br>- Include speculative implementation options and hypothetical future actions that might result in positive effects or implications<br><br>"Opportunities" that apply to a net outcome objective and its associated approach generally exclude:<br><br>- Items that are not clearly linked to net outcome objectives or their associated approaches |
|  | - Items unambiguously unrelated or untransferable to health, community, social, broad environmental, or generic net outcome contexts (e.g., text addressing opportunities related to standalone ecological or resource-orientated objectives)<br><br>"Unstated" should be recorded if the concept is not addressed within the source. "Aggregated (not extracted)" should be recorded if the source presents content not specific to net-outcome aspects of an objective or approach. |
| Threats | "Threats" that apply to an objective and its associated approach:<br>• Are external factors (i.e. not inherent to the net outcome objective and its associated approach)<br>• Are context-specific and relate to its contextual positioning<br>• Include speculative implementation options and hypothetical future actions that might result in negative effects or implications<br><br>"Threats" that apply to a net outcome objective and its associated approach generally exclude:<br>- Items that are not clearly linked to net outcome objectives or their associated approaches<br>- Items unambiguously unrelated or untransferable to health, community, social, broad environmental, or generic net outcome contexts (e.g., text addressing threats related to standalone ecological or resource-orientated objectives)<br><br>"Unstated" should be recorded if the concept is not addressed within the source. "Aggregated (not extracted)" should be recorded if the source presents content not specific to net-outcome aspects of an objective or approach. |

## Appendix III: Data extraction and presentation instructions

The user will upload new source documents as PDFs one at a time for you to process. Process each uploaded PDF document individually as follows:

1. "Scoping review protocol.pdf" - use this project document to define the (broad) concept and scope of relevant net outcome objectives and their associated approaches. These net outcome objectives and their associated approaches are the focus when extracting Implementation principles, Strengths, Weaknesses, Opportunities, and Threats from the new source document PDF.
2. "Data extraction instrument (LLM).docx" [i.e., Appendix II] – use this project document to identify data extraction targets (listed in the left hand column) in new source document PDFs, strictly adhering to any guidance on the concept and scope of each data extraction target (in the right hand column). The specific data extraction targets are Author(s), Publication year, Title, Implementation principles, Strengths, Weaknesses, Opportunities, and Threats.
3. The Author(s), Publication year, and Title should be identified from the first pages of the new source document PDF.
4. If the user specifies one or more specific net outcome objectives that the new source document PDF describes when uploading the PDF into the chat, these specific net outcome objectives and their associated approaches should be a particular focus when extracting Implementation principles, Strengths, Weaknesses, Opportunities, and Threats from the new source document PDF.
5. Extract verbatim text (i.e., direct quotes) from the new source document PDF without any summarising or paraphrasing.
6. Only extract data from the new source document PDF (i.e., do NOT present data extracted from any other sources or project documents).
7. Present data extraction results listed after headers corresponding to each data extraction target (i.e., data extracted for: Author(s), Publication year, Title, Implementation principles, Strengths, Weaknesses, Opportunities, and Threats). The results for each data extraction target should be formatted such that the text can be readily copied and pasted into a single Excel spreadsheet cell.
8. Extracted data related to Implementation principles, Strengths, Weaknesses, Opportunities, and Threats may contain multiple text excerpts from the new source document PDF. Use bullet points to separate distinct text excerpts on new lines. If an extracted excerpt line duplicates the exact meaning of an existing extracted excerpt line for the same data extraction target for the same new source document PDF, omit it (i.e., do not list it twice).

9. If a new source document PDF doesn't contain relevant information for any given data extraction target, state this and provide a brief explanation (in a separate accompanying text narrative).

10. Use a separate accompanying text narrative to flag any: unreadable or corrupted PDFs, errors, uncertain extractions, or to request human review for ambiguous cases.

11. Before presenting outputs to the user, review your output to ensure it adheres to these instructions and that output content is consistent with examples provided in the project document "Data extraction examples.csv".

# Appendix IV: LLM data extraction review prompts

Note to readers: emboldened text in later subsections denotes iterative additions and amendments to earlier prompts.

## *Data extraction review LLM prompt (batches 1-2)*

Your role is to review and check data that has been extracted from source pdfs into the project document spreadsheet titled "Extraction_check_Claude.csv".

After a title row, each row of the spreadsheet titled "Extraction_check_Claude.csv" corresponds to a source pdf whose title is listed in the column titled "Source filename". Source pdfs are uploaded with this prompt. The scope of data extraction is defined by the project document pdf titled "Scoping review protocol.pdf" and the final data extraction instrument described in the word document titled "template-extraction-updated.docx" [i.e., Appendix V].

Following the instructions above, review and check data that has been extracted from source pdfs into the spreadsheet titled "Extraction_check_Claude.csv". To do this, you must compare the content of each uploaded source pdf with the corresponding row on the spreadsheet, considering whether source pdf content has been extracted correctly according to the "Scoping review protocol.pdf" and "template-extraction-updated.docx" [i.e., Appendix V].

Provide separate feedback for each source pdf, using the column titles in the spreadsheet titled "Extraction_check_Claude.csv" to order any comments.  Focus your comments on feeding back unambiguous data extraction errors or omissions. For omissions, specify the excerpt text from the source pdf that is missing.

## *Data extraction review LLM prompt (batch 3)*

Role

Your role is to review and check data that has been extracted from source pdfs into the project document spreadsheet titled "Extraction_check_Claude**_2**.csv".

Background

After a title row, each row of the spreadsheet titled "Extraction_check_Claude**_2**.csv" corresponds to a source pdf whose title is listed in the column titled "Source filename". Source pdfs are uploaded with this prompt. The scope of data extraction is defined by the project document pdf titled "Scoping review protocol.pdf" and the final data extraction instrument described in the word document titled "template-extraction-updated.docx" [i.e., Appendix V].

Task

**Using the role and background** above, review and check data that **I have previously** extracted from source pdfs into the spreadsheet titled "Extraction_check_Claude_**2**.csv". To do this, you must compare the content of each uploaded source pdf with the corresponding row on the spreadsheet, considering whether source pdf content has been extracted correctly according to the "Scoping review protocol.pdf" and "template-extraction-updated.docx" [i.e., Appendix V].

Provide separate feedback for each source pdf, using the column titles in the spreadsheet titled "Extraction_check_Claude_**2**.csv" to order any comments. Focus your comments on feeding back unambiguous data extraction errors or omissions. For omissions, specify the excerpt text from the source pdf that is missing. **Remember that you must compare the content of each source pdf with its corresponding spreadsheet row and your comments and corrections must be drawn solely from the relevant source pdf corresponding to each spreadsheet row.**

*Data extraction review LLM prompt (batch 4)*

Role

Your role is to review and check data that has been extracted from source pdfs into the project document spreadsheet titled "Extraction_check_Claude_2.csv".

Background

After a title row, each row of the spreadsheet titled "Extraction_check_Claude_2.csv" corresponds to a source pdf whose title is listed in the column titled "Source filename". Source pdfs are uploaded with this prompt. The scope of data extraction is defined by the project document pdf titled "Scoping review protocol.pdf" and the final data extraction instrument described in the word document titled "template-extraction-updated.docx" [i.e., Appendix V].

Task

Using the role and background above, review and check data that I have previously extracted from source pdfs into the spreadsheet titled "Extraction_check_Claude_2.csv". To do this, you must compare the content of each uploaded source pdf with the corresponding row on the spreadsheet, considering whether source pdf content has been extracted correctly according to the "Scoping review protocol.pdf" and "template-extraction-updated.docx" **guidelines** [i.e., Appendix V].

Provide separate feedback for each source pdf, using the column titles in the spreadsheet titled "Extraction_check_Claude_2.csv" to order any comments. Focus your comments on feeding back unambiguous data extraction errors or omissions. **For omissions and if proposing additions, specify the excerpt text from the source pdf that is missing.**

**Remember that you must compare the content of each source pdf with the content in its corresponding "Extraction_check_Claude_2.csv" spreadsheet row (i.e., after matching the title of the source pdf with the filename in the column titled "Source filename") – verbatim direct quotes must only originate from its one corresponding source pdf. Your comments, corrections and additions must be drawn solely from the relevant source pdf that corresponds to each spreadsheet row (i.e., do not propose corrections or additional excerpts from any other sources).**

# Appendix V: Scoping review final data extraction instrument (annotated)

| Evidence source details and characteristics | |
|---|---|
| Source # | |
| Citation details | Author(s) *(Vancouver reference style)* <br><br> Publication date <br><br> Title <br><br> Journal (or other publication source) <br><br> Volume <br><br> Issue <br><br> Pages <br><br> Author (or other) keywords (Addition to original protocol) |
| Source perspective (Amendment to original protocol) | *Final categories:* <br><br> *Assessment, Conservation, Corporate, Design and development, Extractive industries, Philosophy, Public health, Regulation, Security, Spatial planning* |
| Country of origin | |
| Type of evidence source (document type) | *Final categories:* <br><br> *Audiovisual material, Book, Book Chapter, Gray Literature, Journal Article, Thesis* |
| Document aims / purpose | |
| Type of net-outcome objective (Addition to original protocol) | *Final categories:* <br><br> *1) A standalone health or wellbeing net-outcome objective (e.g., HNG)* <br><br> *2) A net-outcome objective with a subsidiary/secondary health or wellbeing net-outcome objective (e.g., BNG+wellbeing NWO)* <br><br> *3) A composite/multi-outcome net-outcome objective incorporating a specific health or wellbeing net-outcome objective* |
|   | 4) A standalone social or community net-outcome objective (e.g., SNG, CNG) |
|   | 5) A net-outcome objective with a subsidiary/secondary social or community net-outcome objective (e.g., BNG+social NWO) |
|   | 6) A composite/multi-outcome net-outcome objective relevant to or encompassing health (e.g., ENG, socioecological net gain) |
|   | 7) A generic net-outcome objective (derived from a non-health objective) (e.g., NG/NNL) |
|   | 8) Universal features of or considerations for net-outcome objectives (e.g., equity, offsetting, MH) |
| Health net-outcome objective, target, aim or goal | *Specified 'net' objective, extracted short-form, wherever stated as such* *(extracted for all sources that include a specific health net-outcome objective and any other (e.g., biodiversity or nature-orientated) primary net-outcome objectives that present a secondary community, health, social, or wellbeing objective)* |
| Derived health net-outcome objective (Addition to original protocol) | *Coded shorthand for target and outcome based on field above and source content* *B=Biodiversity, H=Health, N=Nature, S=Social, W=Wellbeing* *DNH+=Do No Harm or better off, NG=Net Gain, NWO=No Worse Off, NWO+=No Worse Off or better off* |
| Primary net-outcome objective, target, aim or goal | *Specified 'net' objective, extracted short-form, wherever stated as such* *(extracted for all sources)* |
| Derived primary net-outcome objective (Addition to original protocol) | *Coded shorthand for target and outcome based on field above and source content* *\*=multi-objective, B=Biodiversity, C=Community, E=Environmental, H=Health, N=Nature, S=Social, W=Wellbeing* *NG=Net Gain, NNL=No Net Loss, NO=Net Outcome* |

|  | *Note: outcome precedes target (in brackets) for generic net-outcome objectives and universal features (refer to field: Type of net outcome)* |
|---|---|
| Country / countries of application | *The nation the objective (or related findings) apply to / are contextualized by, if not universal (noting that sources are considered to have universal applicability by default)* |
| Scale(s) of application (Addition to original protocol) | *Considering both:*<br>*1) the scale(s) the source applies the objective to*<br>*2) the scale(s) the source considers when exploring the objective*<br><br>*Final categories:*<br>*Building*<br>*Project*<br>*City*<br>*Policy (local/regional)*<br>*Policy (national)*<br>*Policy (international)*<br>*Organisation*<br>*System* |
| **Details/results extracted from source of evidence** | |
| Net-objective and approach: characteristics (Amendment to original protocol) | |
| Rationale(s) for net-outcome objective(s) (Amendment to original protocol) | *Rationale for the net-outcome objective (specifically), if stated*<br>*Rationale for the broader objective, if relevant and if no rationale is given for the net-outcome objective specifically*<br>*For net-outcome objectives that present secondary community, health, social, or wellbeing objectives, multiple objectives may be listed on separate lines* |
| Description(s) of net-outcome objective(s) (Amendment to original protocol) | *Description / elaboration of the net-outcome objective that data extraction (subsequent fields) focuses on* <br><br> *For net-outcome objectives that present secondary community, health, social, or wellbeing objectives (refer to field: Type of net outcome), multiple objectives may be listed on separate lines* <br><br> *The default approach is to focus extraction on community, health, social, or wellbeing objectives when specified (refer to field: Type of net outcome)* |
| Definition of objective's health term(s) | *Definition of "health" and/or health term(s), if given* |
| Net-outcome definition / characterisation and/or metric(s) (Addition to original protocol) | *Includes extraction of summary commentary on composite / generic "net-outcome" measurement (but excludes full indicator breakdowns for compositive objectives, for which only a summary precis and/or selected health / planning indicators are extracted - see below)* <br><br> *Includes key health or wellbeing indicators if a net-outcome objective is specifically applied to health or wellbeing (but excludes such indicators when mentioned more generally without reference to health net outcomes)* <br><br> *If outcome definitions are extracted, extensive / elaborating / further measurement requirements / steps are recorded in the "Implementation principles and/or steps" field* |
| Net-outcome level emphasised (Addition to original protocol) | *The level of operationalisation that <u>extracted source data</u> presents, emphasises or focuses on* <br><br> *Final categories:* <br> *1A Overarching requirement* <br> *1B Overarching requirement (locally defined)* <br> *2 Enhanced process* <br> *3A Assessed outcome* |

|  | *3B Assessed outcome (locally defined)* |
|---|---|
| Specific metric(s) or broader framework(s) (Addition to original protocol) | *Where specified beyond generic indicator themes (e.g., environmental, economic, health) or methodological approaches (e.g., qualitative, quantitative)* |
| Implementation principle(s) and/or steps | *The concept of "implementation principles" of an objective and its associated approach is that implementation principles encompass:*<br><br>• *Principles, rules or requirements that govern implementation*<br>• *Items framed as pre-requisites, absolutes, essentials, necessary steps, imperatives, or propositions*<br>• *Implementation frameworks, "how to" guides, "steps to take", and action lists*<br>• *Descriptions of how an objective is implemented in practice or would be implemented in practice*<br>• *Descriptions of how an objective works or operates in practice or how it would work or operate in practice*<br>• *Descriptions of how an objective is or would be operationalised*<br><br>*If a given implementation principle is framed as a possibility or one of a range of options, it is considered to be within the concept of "opportunities" (and not extracted as an "implementation principle").*<br><br>*The concept of "implementation principles" of an objective and its associated approach is that implementation principles generally exclude:*<br>*1) Items that are not clearly linked to net outcome objectives or approaches (e.g., excluding principles related to broader / multi-outcome objectives, such as regenerative design and development, unless specific "net" principles are disaggregated\*)* |
| | *2) Items unambiguously unrelated or untransferable to health, community, social, broad environmental, or generic net outcome contexts (e.g., text addressing standalone ecological or resource-orientated implementation principles)* <br><br> *\*If, however, a broader objective is defined solely by explicit net-outcomes (such as the "positive development" objective), its principles are included)* <br><br> *\*If, however, the source was categorised as elaborating universal principles / implementation steps (i.e., refer to field: Type of net outcome = 8), its principles are included* |
| Objective and approach: contextual positioning and effects (SWOT) <br><br> (Amendment to original protocol) | |
| Positive effects or implications (Strengths) | *The concept of "strengths" of an objective and its associated approach is that strengths:* <br><br> • *Are internal factors (i.e. inherent to the objective and its associated approach)* <br> • *Are positive effects or implications* <br> • *Include observed, demonstrated, or proven positive effects or implications* <br> • *Include advantages versus the status quo or an existing comparator* <br><br> *The concept of "strengths" of an objective and its associated approach is that strengths generally exclude:* <br><br> *1) Items that are not clearly linked to net outcome objectives or approaches* <br><br> *2) Items unambiguously unrelated or untransferable to health, community, social, broad environmental, or generic net outcome contexts (e.g., text addressing strengths related to standalone ecological or resource-orientated objectives)* |
| | *Extraction for composite and generic net-outcome objectives (refer to field: Type of net outcome) may be limited to items contextualised by references to "net" phrases or concepts).* <br><br> *The field records "Unstated" or "Aggregated" if the concept is not addressed within the source or the source presents content not specific to net-outcome aspects of an objective or approach, respectively.* |
| Negative effects or implications (Weaknesses) | *The concept of "weaknesses" of an objective and its associated approach is that weaknesses:* <br> • *Are internal factors (i.e. inherent to the objective and its associated approach)* <br> • *Are negative effects or implications* <br> • *Include observed, demonstrated, or proven negative effects or implications* <br> • *Include disadvantages versus the status quo or an existing comparator* <br><br> *The concept of "weaknesses" of an objective and its associated approach is that weaknesses generally exclude:* <br> *1) Items that are not clearly linked to net outcome objectives or approaches* <br> *2) Items unambiguously unrelated or untransferable to health, community, social, broad environmental, or generic net outcome contexts (e.g., text addressing weaknesses related to standalone ecological or resource-orientated objectives)* <br><br> *Extraction for composite and generic net-outcome objectives (refer to field: Type of net outcome) may be limited to items contextualised by references to "net" phrases or concepts).* |
| | *The field records "Unstated" or "Aggregated" if the concept is not addressed within the source or the source presents content not specific to net-outcome aspects of an objective or approach, respectively.* |
| Implementation opportunities (Opportunities) | *The concept of "opportunities" that apply to an objective and its associated approach is that opportunities:*<br><br>• *Are external factors (i.e. not inherent to the objective and its associated approach)*<br>• *Are context-specific and relate to its contextual positioning*<br>• *Include speculative implementation options and hypothetical future actions that might result in positive effects or implications*<br><br>*The concept of "opportunities" that apply to an objective and its associated approach is that opportunities generally exclude:*<br><br>*1) Items that are not clearly linked to net outcome objectives or approaches*<br><br>*2) Items unambiguously unrelated or untransferable to health, community, social, broad environmental, or generic net outcome contexts (e.g., text addressing opportunities related to standalone ecological or resource-orientated objectives)*<br><br>*Extraction for composite and generic net-outcome objectives (refer to field: Type of net outcome) may be limited to items contextualised by references to "net" phrases or concepts).*<br><br>*The field records "Unstated" or "Aggregated" if the concept is not addressed within the source or the source presents content not specific to net-outcome aspects of an objective or approach, respectively.* |
| Implementation challenges (Threats) | *The concept of "threats" that apply to an objective and its associated approach is that threats:* |
|  | - *Are external factors (i.e. not inherent to the objective and its associated approach)*<br>- *Are context-specific and relate to its contextual positioning*<br>- *Include speculative implementation options and hypothetical future actions that might result in negative effects or implications*<br><br>*The concept of "threats" that apply to an objective and its associated approach is that threats generally exclude:*<br><br>*1) Items that are not clearly linked to net outcome objectives or approaches*<br><br>*2) Items unambiguously unrelated or untransferable to health, community, social, broad environmental, or generic net outcome contexts (e.g., text addressing threats related to standalone ecological or resource-orientated objectives)*<br><br>*Extraction for composite and generic net-outcome objectives (refer to field: Type of net outcome) may be limited to items contextualised by references to "net" phrases or concepts).*<br><br>*The field records "Unstated" or "Aggregated (not extracted)" if the concept is not addressed within the source or the source presents content not specific to net-outcome aspects of an objective or approach, respectively.* |
| Url<br>(Addition to original protocol) | *Direct weblink to source document (where available)* |

## Ethics approval

This study and the scoping review within which it was conducted followed good scientific conduct and did not require formal ethical approval as the research did not involve human participants or the collection of primary data.

## Competing interests

The authors declare no conflicts of interest.

## Funding

JSE is the recipient of a PhD studentship funded by the UK Health Security Agency (UKHSA). Any views expressed in this protocol are those of the authors and not necessarily those of the UKHSA or the Department of Health and Social Care.

## Author contributions (CRediT)

**James Stewart-Evans:** conceptualization (lead); methodology (lead); investigation (lead); formal analysis (lead); writing – original draft (lead); writing – review and editing (equal). **Emma Wilson:** supervision (supporting); writing – review and editing (equal). **Tessa Langley:** supervision (supporting); writing – review and editing (equal). **Andrew Prayle:** conceptualization (supporting); methodology (supporting); writing – review and editing (equal). **Angela Hands:** supervision (supporting); formal analysis (supporting); writing – review and editing (equal). **Karen Exley:** supervision (supporting); review and editing (equal). **Jo Leonardi-Bee:** supervision (lead); writing – review and editing (equal). plcem

## Review registration number

Open Science Framework https://osf.io/4dbcm

| Evidence source # | Data extraction approach | LLM process order | Relevant excerpts | Misclassified excerpts | Irrelevant excerpts | New excerpts | Ineligible excerpts |
|---|---|---|---|---|---|---|---|
| 1 | Human (baseline) | | 24 | N/A | N/A | N/A | No |
| | Claude LLM (extended protocol) | 1 | 3 | 3 | 1 | 1 | No |
| | Claude LLM (protocol V1.1) | 12 | 0 | 1 | 6 | 1 | No |
| 2 | Human (baseline) | | 16 | N/A | N/A | N/A | No |
| | Claude LLM (extended protocol) | 2 | 2 | 1 | 4 | 0 | Yes |
| | Claude LLM (protocol V1.1) | 13 | 3 | 1 | 4 | 1 | No |
| 3 | Human (baseline) | | 15 | N/A | N/A | N/A | No |
| | Claude LLM (extended protocol) | 3 | 5 | 3 | 2 | 0 | Yes |
| | Claude LLM (protocol V1.1) | 14 | 1 | 2 | 5 | 0 | Yes |
| 4 | Human (baseline) | | 14 | N/A | N/A | N/A | No |
| | Claude LLM (extended protocol) | 4 | 2 | 6 | 0 | 2 | No |
| | Claude LLM (protocol V1.1) | 15 | 3 | 3 | 1 | 3 | No |
| 5 | Human (baseline) | | 3 | N/A | N/A | N/A | No |
| | Claude LLM (extended protocol) | 5 | 2 | 2 | 6 | 0 | No |
| | Claude LLM (protocol V1.1) | 16 | 0 | 2 | 6 | 0 | No |
| 6 | Human (baseline) | | 8 | N/A | N/A | N/A | No |
| | Claude LLM (extended protocol) | 6 | 5 | 2 | 0 | 4 | No |
| | Claude LLM (protocol V1.1) | 17 | 6 | 4 | 0 | 6 | No |
| 7 | Human (baseline) | | 39 | N/A | N/A | N/A | No |
| | Claude LLM (extended protocol) | 7 | 2 | 3 | 0 | 0 | No |
| | Claude LLM (protocol V1.1) | 18 | 4 | 2 | 0 | 0 | No |
| 8 | Human (baseline) | | 21 | N/A | N/A | N/A | No |
| | Claude LLM (extended protocol) | 8 | 9 | 4 | 5 | 0 | No |
| | Claude LLM (protocol V1.1) | 19 | 3 | 2 | 2 | 0 | No |
| 9 | Human (baseline) | | 64 | N/A | N/A | N/A | No |
| | Claude LLM (extended protocol) | 9 | 16 | 6 | 1 | 0 | No |
| | Claude LLM (protocol V1.0) | 11 | 4 | 4 | 2 | 0 | No |
| 10* | Human (baseline) | | 2 | N/A | N/A | N/A | No |
| | Claude LLM (extended protocol) | 10 | 1 | 4 | 4 | 3 | No |
| | Claude LLM (protocol V1.1) | 20 | 2 | 3 | 1 | 2 | No |
| All | Human (baseline) | | 206 | N/A | N/A | N/A | No |
| | Claude LLM (extended protocol) | | 47 | 34 | 23 | 10 | Yes (2 of 10) |
| | Claude LLM (protocol) | | 26 | 24 | 27 | 13 | Yes (1 of 10) |

| Data item | Data extraction method | True positives | True negatives | False positives | False negatives | Accuracy (TP+TN)/(TP+TN+FP+FN) | Precision TP/(TP+FP) | Recall TP/(TP+FN) | F1 score 2*(P*R)/(P+R) |
|---|---|---|---|---|---|---|---|---|---|
| Author(s) | Claude LLM (extended protocol) | 10 | 0 | 0 | 0 | 100% | 100% | 100% | 100% |
|  | Claude LLM (protocol) | 10 | 0 | 0 | 0 | 100% | 100% | 100% | 100% |
| Publication date | Claude LLM (extended protocol) | 10 | 0 | 0 | 0 | 100% | 100% | 100% | 100% |
|  | Claude LLM (protocol) | 10 | 0 | 0 | 0 | 100% | 100% | 100% | 100% |
| Title | Claude LLM (extended protocol) | 10 | 0 | 0 | 0 | 100% | 100% | 100% | 100% |
|  | Claude LLM (protocol) | 5 | 0 | 0 | 5 | 50.0% | 100% | 50.0% | 66.7% |
| *All citation data items (n=3 items)* | *Claude LLM (extended protocol)* | *30* | *0* | *0* | *0* | *100%* | *100%* | *100%* | *100%* |
|  | *Claude LLM (protocol)* | *25* | *0* | *0* | *5* | *83.3%* | *100%* | *83.3%* | *90.9%* |
| Implementation principles | Claude LLM (extended protocol) | 21 | 0 | 1 | 83 | 20.0% | 95.5% | 20.2% | 33.3% |
|  | Claude LLM (protocol) | 5 | 0 | 1 | 105 | 4.5% | 83.3% | 4.5% | 8.6% |
| Strengths | Claude LLM (extended protocol) | 4 | 0 | 0 | 32 | 11.1% | 100% | 11.1% | 20.0% |
|  | Claude LLM (protocol) | 5 | 0 | 2 | 28 | 14.3% | 71.4% | 15.2% | 25.0% |
| Weaknesses | Claude LLM (extended protocol) | 5 | 1 | 1 | 26 | 18.2% | 83.3% | 16.1% | 27.0% |
|  | Claude LLM (protocol) | 3 | 3 | 0 | 23 | 20.7% | 100% | 11.5% | 20.7% |
| Opportunities | Claude LLM (extended protocol) | 6 | 0 | 1 | 51 | 10.3% | 85.7% | 10.5% | 18.8% |
|  | Claude LLM (protocol) | 3 | 0 | 1 | 55 | 5.1% | 75.0% | 5.2% | 9.7% |
| Threats | Claude LLM (extended protocol) | 4 | 1 | 0 | 29 | 14.7% | 100% | 12.1% | 21.6% |
|  | Claude LLM (protocol) | 7 | 0 | 1 | 28 | 19.4% | 87.5% | 20.0% | 32.6% |
| *All key findings data items (n=5 items)* | *Claude LLM (extended protocol)* | *40* | *2* | *3* | *221* | *15.8%* | *93.0%* | *15.3%* | *26.3%* |
|  | *Claude LLM (protocol)* | *23* | *3* | *5* | *239* | *9.6%* | *82.1%* | *8.8%* | *15.9%* |
| *All data items above (micro) (n=8 items)* | *Claude LLM (extended protocol)* | *70* | *2* | *3* | *221* | *24.3%* | *95.9%* | *24.1%* | *38.5%* |
|  | *Claude LLM (protocol)* | *48* | *3* | *5* | *244* | *17.0%* | *90.6%* | *16.4%* | *27.8%* |
| *All data items above (macro) (n=8 items)* | *Claude LLM (extended protocol)* |  |  |  |  | *46.8%* | *95.6%* | *46.3%* | *52.6%* |
|  | *Claude LLM (protocol)* |  |  |  |  | *39.3%* | *89.7%* | *38.3%* | *45.4%* |

| Evidence source # | Batch | LLM citation corrections | Added value? | LLM additional excerpts | Added value? | Ineligible excerpts |
|---|---|---|---|---|---|---|
| 1 | 1st | 2 | 1 | 9 | 1 | Yes |
| 2 | 1st | 2 | 0 | 6 | 0 | No |
| 3 | 1st | 2 | 0 | 6 | 0 | Yes |
| 4 | 1st | 4 | 0 | 6 | 0 | No |
| 5 | 1st | 2 | 0 | 8 | 5 | Yes |
| 6 | 2nd | 1 | 1 | 2 | 2 | No |
| 7 | 2nd | 0 | 0 | 0 | 0 | No |
| 8 | 2nd | 2 | 2 | 0 | 0 | No |
| 9 | 2nd | 0 | 0 | 1 | 0 | No |
| 10 | 2nd | 0 | 0 | 0 | 0 | No |
| 1 to 5 | 1st | 12 | 1 | 35 | 6 | Yes (3 of 5) |
| 6 to 10 | 2nd | 3 | 3 | 3 | 2 | No (0 of 5) |
| All | All | 15 | 4 | 38 | 8 | Yes (3 of 10) |

| Source # | Batch | Publication year error | Objective type error | Data extraction target error | Ineligible source | Random text inclusion |
|---|---|---|---|---|---|---|
| 1 | 3rd | Undetected | Undetected | N/A | Undetected | Undetected |
| 2 | 3rd | Undetected | Undetected | Undetected | Undetected | N/A |
| 3 | 3rd | Detected | Undetected | Undetected | Undetected | N/A |
| 4 | 3rd | Undetected | Undetected | Undetected | Undetected | Undetected |
| 5 | 3rd | Undetected | Undetected | N/A | Undetected | N/A |
| 6 | 4th | Undetected | N/A | Undetected | Undetected | Detected |
| 7 | 4th | Undetected | Undetected | Undetected | Undetected | Undetected |
| 8 | 4th | Undetected | Undetected | Undetected | Undetected | N/A |
| 9 | 4th | Undetected | N/A | Undetected | Undetected | N/A |
| 10 | 4th | Undetected | N/A | Undetected | Undetected | N/A |
| 1 to 5 | 3rd | 1 of 5 | 0 of 5 | 0 of 3 | 0 of 5 | 0 of 2 |
| 6 to 10 | 4th | 0 of 5 | 0 of 2 | 0 of 5 | 0 of 5 | 1 of 2 |
| **All** | **All** | 1 of 10 | 0 of 7 | 0 of 8 | 0 of 10 | 1 of 4 |